%% file: main.tex
\documentclass[runningheads]{llncs}
\usepackage[T1]{fontenc}
\usepackage{graphicx}
\usepackage{subcaption}
\usepackage{booktabs}
\usepackage{multirow}
\usepackage{amssymb}
\usepackage{amsmath}
\usepackage{mathrsfs}
\usepackage{wrapfig,booktabs}
\usepackage[font=footnotesize,labelfont=bf]{caption}
\usepackage[english]{babel}
\usepackage{hyperref}
\usepackage{color}

\begin{document}
\title{
ProReco: A Process Discovery \\ Recommender System
% \thanks{Supported by organization x.}
}
%
%\titlerunning{Abbreviated paper title}
% If the paper title is too long for the running head, you can set
% an abbreviated paper title here
%
\author{Tsung-Hao Huang\inst{1}\orcidID{0000-0002-3011-9999} \and
Tarek Junied\inst{2}\orcidID{0000-0001-9318-9276}\and \\
Marco Pegoraro\inst{1}\orcidID{0000-0002-8997-7517} \and \\
Wil M. P. van der Aalst\inst{1}\orcidID{0000-0002-0955-6940}}
\authorrunning{T. Huang et al.}
% First names are abbreviated in the running head.
% If there are more than two authors, 'et al.' is used.
%
\institute{Process and Data Science (PADS), RWTH Aachen University, Aachen, Germany \\
\email{\{tsunghao.huang,pegoraro,wvdaalst\}@pads.rwth-aachen.de}\\
\url{http://www.pads.rwth-aachen.de/} \and
RWTH Aachen University, Aachen, Germany\\
\email{tarekjunied@icloud.com}
}
\maketitle              % typeset the header of the contribution
\begin{abstract}
Process discovery aims to automatically derive process models from historical execution data (event logs). 
While various process discovery algorithms have been proposed in the last 25 years, there is no consensus on a dominating discovery algorithm. 
Selecting the most suitable discovery algorithm remains a challenge due to competing quality measures and diverse user requirements. 
Manually selecting the most suitable process discovery algorithm from a range of options for a given event log is a time-consuming and error-prone task. 
This paper introduces \texttt{ProReco}, a \textbf{Pro}cess discovery \textbf{Reco}mmender system designed to recommend the most appropriate algorithm based on user preferences and event log characteristics. 
ProReco incorporates state-of-the-art discovery algorithms, extends the feature pools from previous work, and utilizes eXplainable AI (XAI) techniques to provide explanations for its recommendations. 

\keywords{Process Mining \and Process Discovery \and Recommender System \and Explainable Recommendations \and Explainable AI.}
\end{abstract}

\input{sections/intro}

\input{sections/preliminaries}

\input{sections/ProReco}

\input{sections/conclusion}

\vspace{-1em}
\subsubsection{Acknowledgements.}
We thank the Alexander von Humboldt (AvH) Stiftung for
supporting our research.

\bibliographystyle{splncs04}

\typeout{}
% \vspace{-1em}
\bibliography{myrefs}

\end{document}

%% file: sections/intro.tex
\section{Introduction}\label{sec:inro}

Process discovery~\cite{Aalst16PMbook} is a discipline that aims to automatically obtain formal representation through models of the operating mechanisms in a process. The input of such methods is a collection of data related to the historical execution of a process, often in the form of discrete \emph{events}. 
Discovery algorithms read events and their \emph{attributes} from a dataset (often called an \emph{event log}), and output a process model, to provide a representation as close as possible to the real process operations.

Since the inception of the discipline in the early 2000s, many discovery algorithms have been proposed~\cite{DBLP:books/sp/22/Augusto0022}, as well as numerous metrics to assess their desirability and quality. 
Nevertheless, the systematic review and benchmark~\cite{AugustoCDRMMMS19PDreviewbenchmark} show no algorithm dominating all other methods in terms of model quality measures. 
% Many discovery approaches use Petri nets as their representation of choice, since it is one of the simplest formalisms that explicitly model concurrency. 
% Other approaches leverage transition systems, process trees, and BPMN diagrams as their output representations. 
Moreover, producing a satisfactory process model is still an open challenge, although there exists extensive literature dedicated to measuring the quality of models obtained through discovery. 
This is because (i) some of the most widely adopted quality measures are competing (i.e., there exist trade-offs between them), and (ii) depending on the final use of the discovered model, different (and sometimes opposite) characteristics are desirable. 
Under such a circumstance, users are left with the task of manually selecting the most prominent process discovery algorithm for the event log at hand. 
The procedure is time-consuming and error-prone even for process mining experts, let alone inexperienced users. 

To address the aforementioned problems and to assist process mining users, previous works~\cite{RibeiroCMS14Recommend,TavaresJD22MetaRecommend} proposed using recommender systems for process discovery. 
% However, the algorithm portfolio and feature extraction are not up-to-date with the recent developments in process discovery technology. 
The approaches~\cite{RibeiroCMS14Recommend,TavaresJD22MetaRecommend} abstract from the actual values of model quality by calculating the final score based on the rankings. 
Also, the approach in~\cite{TavaresJD22MetaRecommend} does not incorporate user preferences for the recommendation, assuming every user wants to maximize all measures simultaneously. 
Lastly, the recommendations offered by both works lack accompanying explanations. 
Intransparent recommendations could hamper the acceptance of a recommender system~\cite{ZhangC20XRecommendation}. 

This paper proposes \texttt{ProReco}, a \textbf{Pro}cess discovery \textbf{Reco}mmender system. 
Given an event log and user preferences regarding model quality measures, \texttt{ProReco} recommends the most appropriate process discovery algorithm tailored to the users' needs. 
Internally, \texttt{ProReco} utilizes machine learning models to predict the values for each quality measure before computing the weighted (user preferences) sum of the final score. 
The scores are then used to rank and recommend the discovery algorithm. 
\texttt{ProReco} not only expands the features pool from previous work but also includes state-of-the-art process discovery algorithms.  
Last but not least, for every recommendation made by \texttt{ProReco}, explanations are available to the user thanks to the incorporation of the eXplainable AI (XAI) technique~\cite{LundbergL17SHAP} in \texttt{ProReco}. 

The remainder of the paper is structured as follows. Section~\ref{sec:prel} illustrates some preliminary notions. Section~\ref{sec:ProReco} describes the components and mechanics of \texttt{ProReco}. Lastly, Section~\ref{sec:conclusion} concludes the paper and indicates directions for future research.

%% file: sections/preliminaries.tex
\section{Preliminary Concepts}\label{sec:prel}
\vspace{-0.5em}
In this section, we introduce the necessary concepts before presenting \texttt{ProReco} in Sec.~\ref{sec:ProReco}. 

\subsubsection{Event log}
\begin{figure}[h!]
    % \vspace{-1.5em}
    \centering
        \begin{subfigure}[b]{.33\linewidth}
            \includegraphics[width=\linewidth]{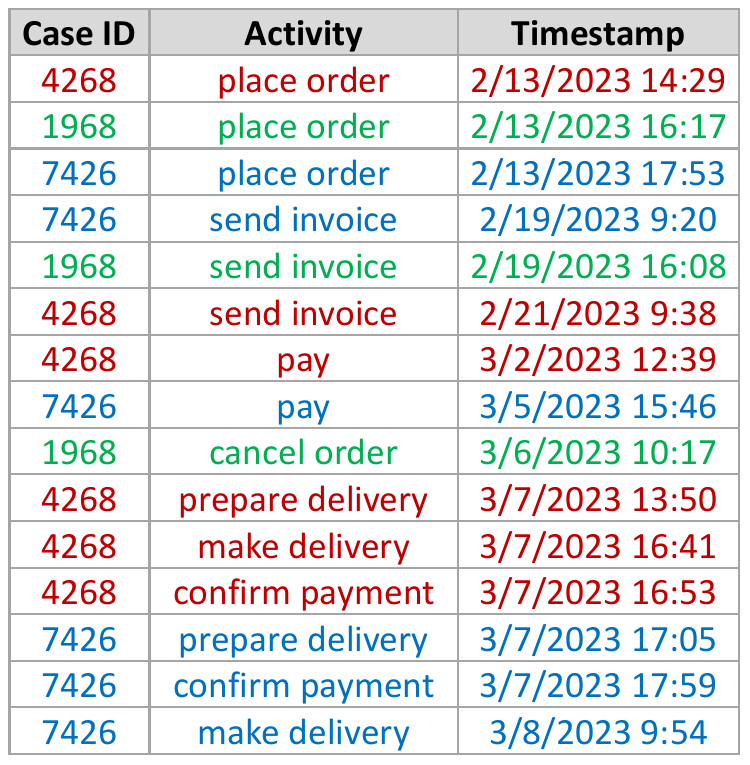}
            \caption{An example event log}\label{fig:preli-log}
        \end{subfigure}
        % \hspace{0.5em} % for more space between subfigures
        \begin{subfigure}[b]{.65\linewidth}
            \includegraphics[width=\linewidth]{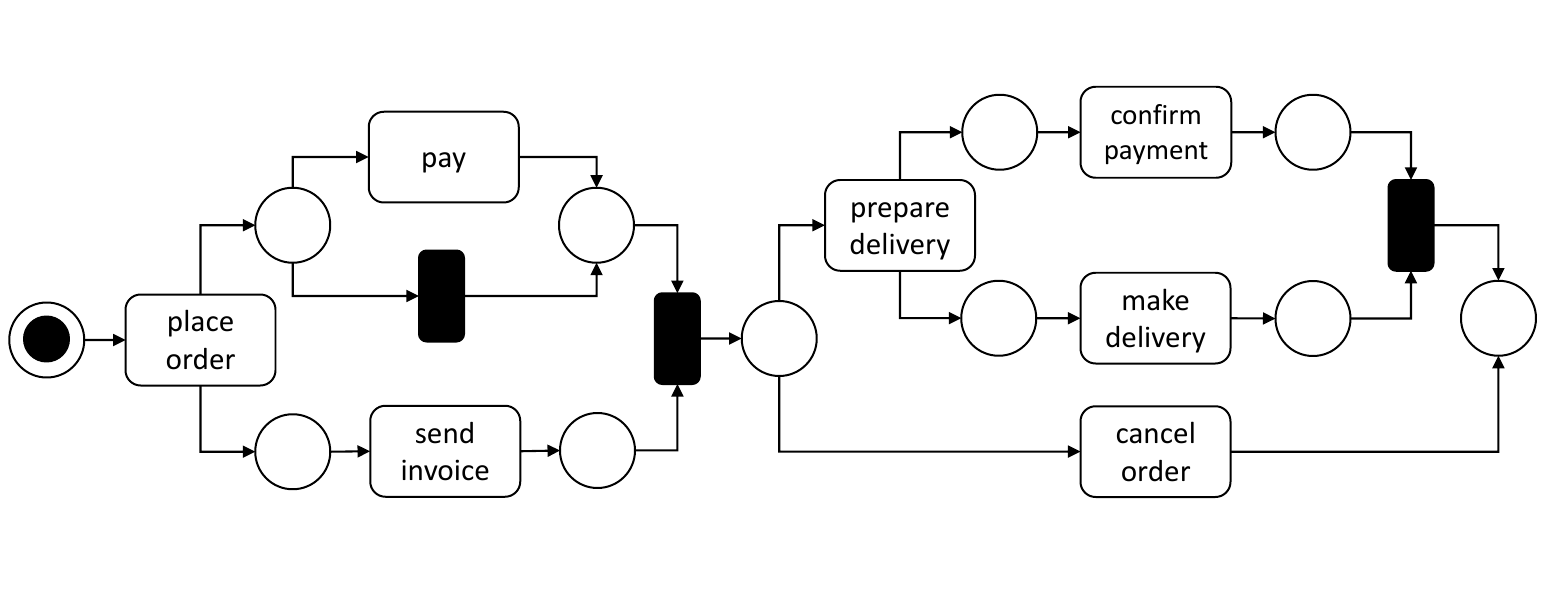}
            \caption{A process model represented using Petri net}\label{fig:preli-model}
        \end{subfigure}
    \caption{An example of an event log and the corresponding process model.}
    \label{fig:preli-log-model}
    \vspace{-1.5em}
\end{figure}

The starting point of process mining is the event log where each event refers to a case (an instance of the process), an activity, and a point in time. 
The existence of these three attributes is the minimal requirement for an event log, whereas more attributes can be recorded and/or extracted. 
Event data can be extracted from various sources such as a database, a transaction log, a business suite/ERP system, etc. 
An event log can be seen as a collection of cases, whereas a case is a trace/sequence of events. 
Fig.~\ref{fig:preli-log} shows a synthetic event log for the purchasing process of an online retail site. 
Each row corresponds to an event. 
% An event is associated with a case (an instance of the process), an activity name, and a timestamp. 
% It's crucial to associate each event with its corresponding case so that process mining tools can compare multiple process executions effectively. 
% Activity names refer to different process steps or status changes in the process. 
% Lastly, the timestamp column indicates when each of the events took place. 

\subsubsection{Process model}

A process model is a structured representation of the activities and their relationships within a business process. 
It plays a crucial role in understanding, analyzing, and improving organizational workflows. 
Various process modeling notations exist such as Petri nets, BPMNs, BPEL models, or UML Activity Diagrams~\cite{Aalst16PMbook}.  
% Many fundamental discovery approaches use Petri nets as their representation of choice while other approaches leverage transition systems, process trees, and BPMN diagrams as the output representations. 
In \texttt{ProReco}, we focus on Petri net since it is one of the simplest formalisms that explicitly model concurrency. 
Moreover, it is trivial to convert process models in other notations into Petri nets.  

% A Petri net is a bipartite model consisting of places (circles), transitions (squares), and directed arcs. 
% Places can hold tokens (black dots) while transitions produce and/or consume tokens. 
% A transition is enabled if each input place contains a token. 
% An enabled transition can fire by consuming one token from each input place and producing a token for each output place. 
% By such a simple mechanism, Petri nets are used to represent and analyze the behavior of complex concurrent systems. 
Fig.~\ref{fig:preli-model} shows the corresponding process model (in the form of a Petri net) for the event log in Fig.~\ref{fig:preli-log}. 
The process starts with the activity \textit{``place order''} followed by the concurrent executions of activity \textit{``pay''} and \textit{``send invoice''}, where activity \textit{``pay''} is optional. 
Then, the process might be either \textit{``cancel''} or followed by a delivery procedure. 

\subsubsection{Process discovery}
Process discovery aims at constructing process models to describe the observed behaviors of information systems from event logs. 
In general, the problem of process discovery can be defined as follows: 
A process discovery algorithm is a function that maps an event log $L$ onto a process model $N$ such that the model $N$ is representative of the behaviors seen in the log $L$. 
Despite the development of process discovery algorithms, manually finding the most appropriate algorithm is a challenging and error-prone task. 
To assist users with identifying the most prominent discovery algorithm, we present \texttt{ProReco} in the next section.

% \begin{table}[]
% \centering
% \caption{}
% \label{tab:event-log}
% \begin{tabular}{|c|c|c|}
% \hline
% Case ID & Activity         & Timestamp       \\ \hline
% 4268    & place order      & 2/13/2023 14:29 \\ \hline
% 1968    & place order      & 2/13/2023 16:17 \\ \hline
% 7426    & place order      & 2/13/2023 17:53 \\ \hline
% 7426    & send invoice     & 2/19/2023 9:20  \\ \hline
% 1968    & send invoice     & 2/19/2023 16:08 \\ \hline
% 4268    & send invoice     & 2/21/2023 9:38  \\ \hline
% 4268    & pay              & 3/2/2023 12:39  \\ \hline
% 7426    & pay              & 3/5/2023 15:46  \\ \hline
% 1968    & cancel order     & 3/6/2023 10:17  \\ \hline
% 4268    & prepare delivery & 3/7/2023 13:50  \\ \hline
% 4268    & make delivery    & 3/7/2023 16:41  \\ \hline
% 4268    & confirm payment  & 3/7/2023 16:53  \\ \hline
% 7426    & prepare delivery & 3/7/2023 17:05  \\ \hline
% 7426    & confirm payment  & 3/7/2023 17:59  \\ \hline
% 7426    & make delivery    & 3/8/2023 9:54   \\ \hline
% \end{tabular}
% \end{table}

%% file: sections/ProReco.tex
\section{ProReco: A Process Discovery Recommender System}\label{sec:ProReco}
The backend of \texttt{ProReco} is developed in Python, to leverage the capabilities of the PM4py\footnote{\url{https://pm4py.fit.fraunhofer.de/}} package. 
The package provides a comprehensive suite of algorithms and tools for process mining. 
The source code for \texttt{ProReco} can be found on a GitHub repository\footnote{\url{https://github.com/TarekJunied/ProReco}}, which provides detailed instruction for installation. 
% The frontend of \texttt{ProReco} is developed using the React\footnote{\url{https://react.dev/}} framework. 
% Lastly, the Petri net viewer 
In the following, we introduce the structure and the main functions of \texttt{ProReco}. 
Additionally, a demo video of \texttt{ProReco} is available\footnote{\url{https://bit.ly/prorecodemo}}. 
\subsection{Structure}\label{subsec:ProReco-structure}
\begin{figure}[h!]
    \vspace{-1.5em}
    \centering
    \includegraphics[width=0.8\linewidth]{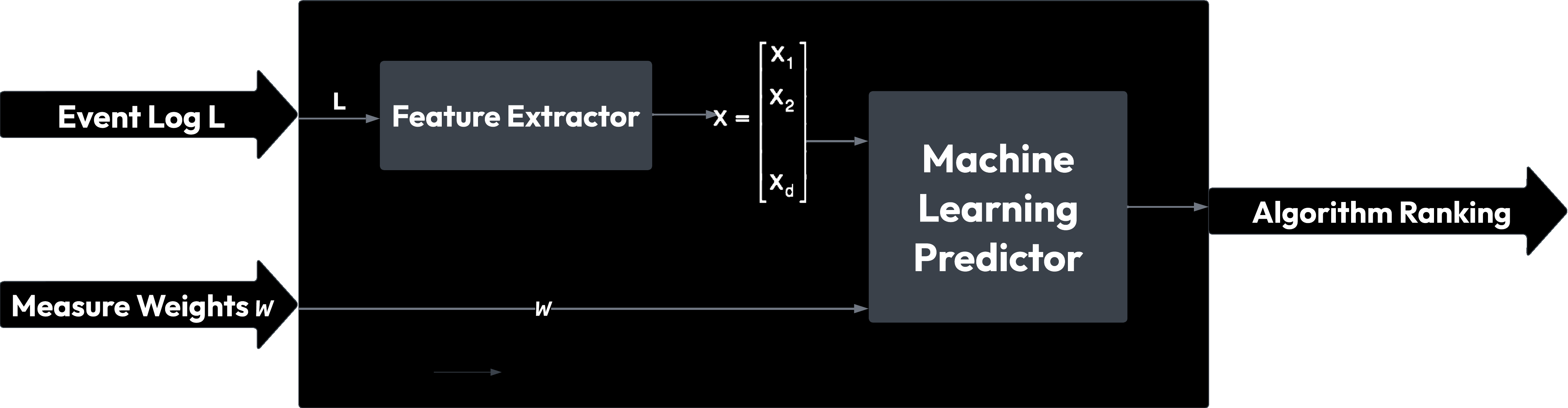}
    \caption{
        General structure of \texttt{ProReco}
    } \label{fig:structure}
    \vspace{-1.5em}
\end{figure}

The overall structure of \texttt{ProReco} is shown in Fig.~\ref{fig:structure}. 
To recommend the most prominent discovery algorithm for event log $L$, \texttt{ProReco} takes a vector of weights $W$ representing the importance of different measures in addition to $L$. 
The weights (within the interval [0,100]) are given by the users and will be used to calculate the final score of the algorithm. 

The output of \texttt{ProReco} is a ranking for each discovery algorithm in our portfolio, as well as the corresponding score calculated based on the weighted sum of the quality measures. 
% The higher the score, the better the algorithm is toward the users' preferences. 
The higher the score, the better the algorithm adapts to the users' preferences. 
In the following, we briefly describe the target quality dimensions used in \texttt{ProReco}.  

As we aim to quantify the most common quality measures of a process model, the four primary quality measures~\cite{Aalst16PMbook} (\emph{fitness}, \emph{simplicity}, \emph{precision}, and \emph{generalization}) are used. 
%A model with good fitness allows for behaviors seen in the log.
A model with good fitness represents (and can replay) behavior seen in the log. 
The simplicity dimension refers to the complexity of the model. 
%In the context of process mining, this means the simpler the model the better as long as it can explain the behaviors seen in the log. 
In the context of process mining, this means that a simpler model is advantageous, as long as it can explain the behaviors seen in the log. 
A precise process model does not allow too much unseen behavior, as it is trivial to create a model that allows any behavior (the flower model~\cite{Aalst16PMbook}). 
%Lastly, a generalized model doesn't restrict the behaviors to only the ones in the log. 
Lastly, a model with good generalization can represent behavior unseen in the event log. 
Since the four quality dimensions compete with each other, a single ideal model often does not exist~\cite{Aalst16PMbook}. 
The ideal model highly depends on the use case of the users. 
This motivates the use of weights to incorporate the user preferences w.r.t. the importance of the four quality measures. 

Next, we introduce each component (the feature extractor and the machine learning predictor) in more detail. 

\subsubsection{Feature Extractor}
Based on previous work~\cite{RibeiroCMS14Recommend,TavaresJD22MetaRecommend,Zandkarimi2021Fig4PM}, we extract an initial pool of various features. 
Moreover, the initial pool is filtered considering two criteria. 
First, we remove the computationally expensive features. 
As efficiency is one of the motivations for developing such a recommender system, using features that are expensive to compute does counteract the benefit. 
% Therefore, several graph-based features from~\cite{Zandkarimi2021Fig4PM} are removed. 
Second, we remove redundant features. 
Features are considered redundant if there is already another feature representing the same concept. 
For instance, the feature representing the number of trace variants is implemented as \texttt{n\_unique\_traces} in~\cite{TavaresJD22MetaRecommend} and as “Number of distinct traces” in~\cite{RibeiroCMS14Recommend}. 
These redundancies lead to higher execution time and adversely affect the performance of some machine learning models. 
Thus, we remove such redundancies using the Pearson correlation coefficient. 
Lastly, we add ten Directly-Follows Graph (DFG)- and footprint matrix-based features. 
In the end, 162 features were extracted. 
The introduction to all features is out of scope. 
The corresponding function (called \textit{Featurer}) providing insight for all features is available in \texttt{ProReco} and introduced in Sec.~\ref{subsec:ProReco-functions} in more detail. 

\subsubsection{Machine Learning Predictor}
\begin{figure}[h!]
    \vspace{-2em}
    \centering
        \begin{subfigure}[b]{.9\linewidth}
            \includegraphics[width=\linewidth]{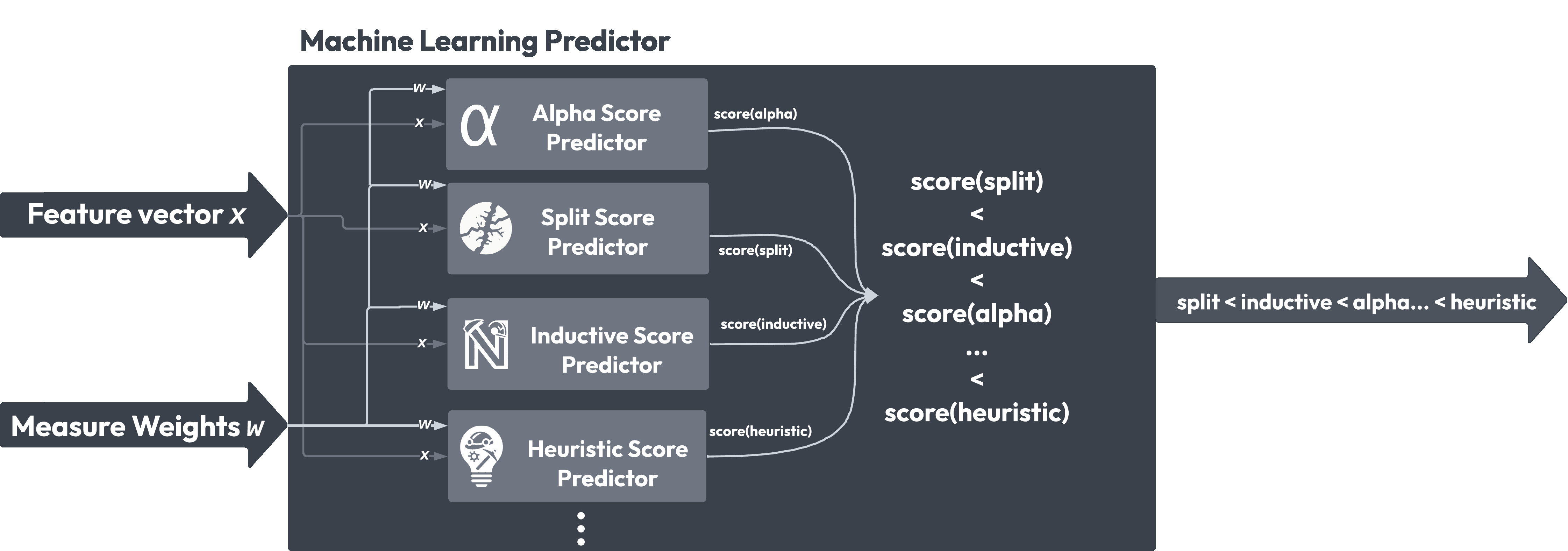}
            \caption{
                Machine learning predictor component.
            } \label{fig:ML-predictor}
        \end{subfigure}
        % \hspace{0.5em} % for more space between subfigures
        \begin{subfigure}[b]{.9\linewidth}
            \includegraphics[width=\linewidth]{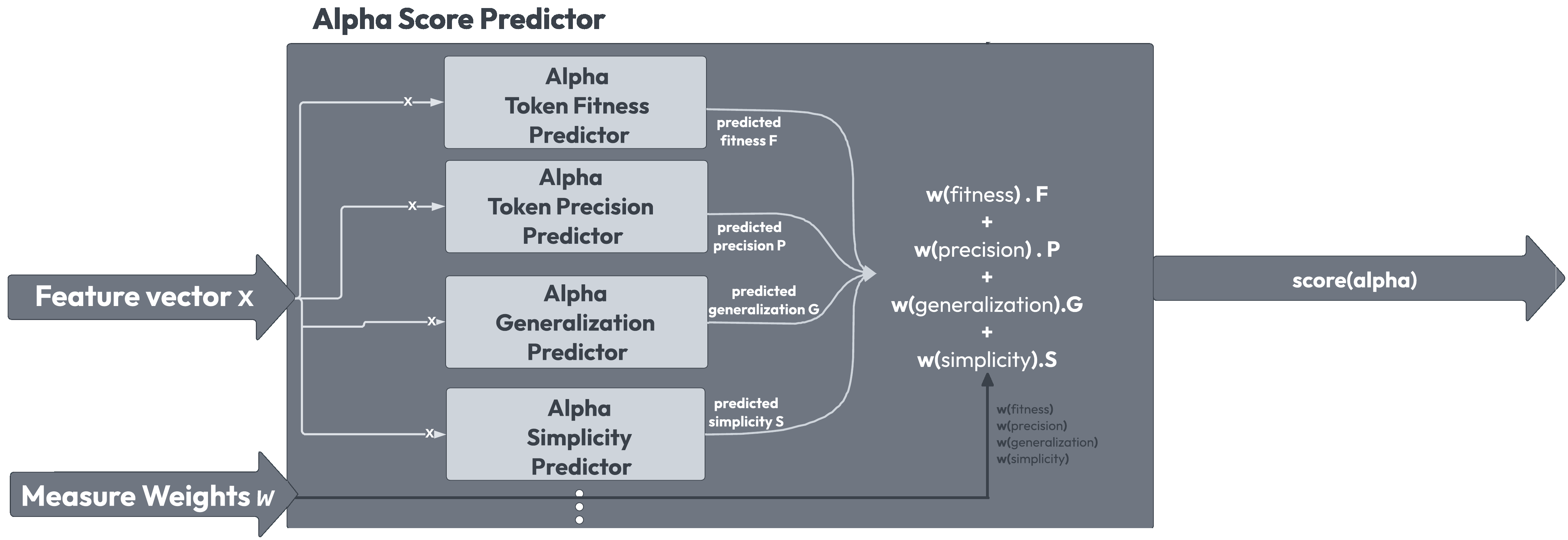}
                \caption{
                    Score predictor for a single algorithm (Alpha Miner as an example)
                } \label{fig:alpha-predictor}
        \end{subfigure}
    \caption{Machine learning predictor and its sub-components: score predictors.}
    \label{fig:ML-predictor-component}
    \vspace{-1.5em}
\end{figure}

As shown in Fig.~\ref{fig:ML-predictor}, the machine learning predictor consists of a score predictor for each algorithm in the algorithm portfolio, which consists of Alpha Miner, Alpha-Plus Miner, Heuristics Miner, Inductive Miner (classic, infrequent, direct), ILP Miner, and Split Miner. 
% As our algorithm portfolio consists of eight discovery algorithms (Alpha Miner, Alpha-Plus Miner, Heuristics Miner, Inductive Miner (classic, infrequent, direct), ILP Miner, and Split Miner), there are 32 machine learning predictors in total. 
% Inside the score predictor for each algorithm, there is a predictor for each measure (fitness, precision, generalization, and simplicity). 
% Therefore, we have $8\times 4=32$ measure predictors in the end.

% These predictors receive as input a feature vector representing an event log, along with the measure weights, to generate a score specific to the algorithm. 
% The score is determined by computing a weighted (based on the provided measure weights) sum of the predicted measure values. 

% \begin{figure}[h!]
%     \centering
%     \includegraphics[width=0.8\linewidth]{figs/ML-predictor.png}
%     \caption{
%         Machine learning predictor component.
%     } \label{fig:ML-predictor}
%     % \vspace{-2em}
% \end{figure}
The structure for an algorithm score predictor is shown in Fig.~\ref{fig:alpha-predictor} using Alpha Miner as an instance. 
The Score Predictor consists of individual predictors for each of the four measures (fitness, precision, generalization, and simplicity). 
Each measure predictor accepts a feature vector derived from the event log as input and forecasts the value of the corresponding measure for the process model that would have been generated based on the provided event log. 
The measure weights $W$ provided by the user are then used for the final computation of the overall score for a discovery algorithm. 
During this computation, each predicted measure value is multiplied by its corresponding measure weight and subsequently aggregated. 

% \begin{figure}[h!]
%     \centering
%     \includegraphics[width=0.8\linewidth]{figs/alpha-predictor.png}
%     \caption{
%         Score predictor for a single algorithm (Alpha Miner as an example)
%     } \label{fig:alpha-predictor}
%     \vspace{-1em}
% \end{figure}

The choice of the predictor to predict the measure values for each algorithm is of little importance here, as it is flexible to switch among different predictors whenever suitable. 
In \texttt{ProReco}, we adopt the \texttt{xgboost}~\cite{ChenG16XGBoost} regressor as an instantiation for the predictor. 
To train the predictors, we included 12 real-life event logs from the 4TU repository\footnote{\url{https://data.4tu.nl/}} and 785 synthetic event logs generated by the \texttt{PLG} tool\footnote{\url{https://plg.processmining.it/}}. 
The data is available for download\footnote{\url{http://bit.ly/allEventLogsProReco}}. 
We used 5-fold cross-validation for each experiment with an 80/20 training/test split.

\subsection{ProReco's Functions}\label{subsec:ProReco-functions}
In this section, we introduce the main functions of \texttt{ProReco}. 
% Additionally, a demo video of \texttt{ProReco} is available\footnote{\url{https://bit.ly/prorecodemo}}. 

\vspace{-0.5em}
\subsubsection{Recommendation}
As a recommender system for process discovery, \texttt{ProReco} recommends the most prominent algorithm for the user according to the predicted weighted sum of the four quality measures discussed in Sec.~\ref{subsec:ProReco-structure}. 
The inputs are an event log $L$ and the user preferences w.r.t. measure weights $W$.

To initiate the recommendation, users have to upload an event log (\texttt{.xes} format) as input. 
Then, they are redirected to a page where they have to provide the weights for each of the four quality measures. 
Once the measure weights are submitted, users are redirected to the recommendation page, where a ranking of the algorithm portfolio, the score for each algorithm, and the predicted measure values for each algorithm and measure are available. 
Additionally, \texttt{ProReco} provides users the possibility to mine a process model with the recommended process discovery algorithms. 
The discovered process model is then displayed through an interactive Petri net viewer. 

\vspace{-0.5em}
\subsubsection{Feature Insights}
\begin{figure}[h!]
    % \vspace{-1.5em}
    \centering
    \includegraphics[width=0.9\linewidth]{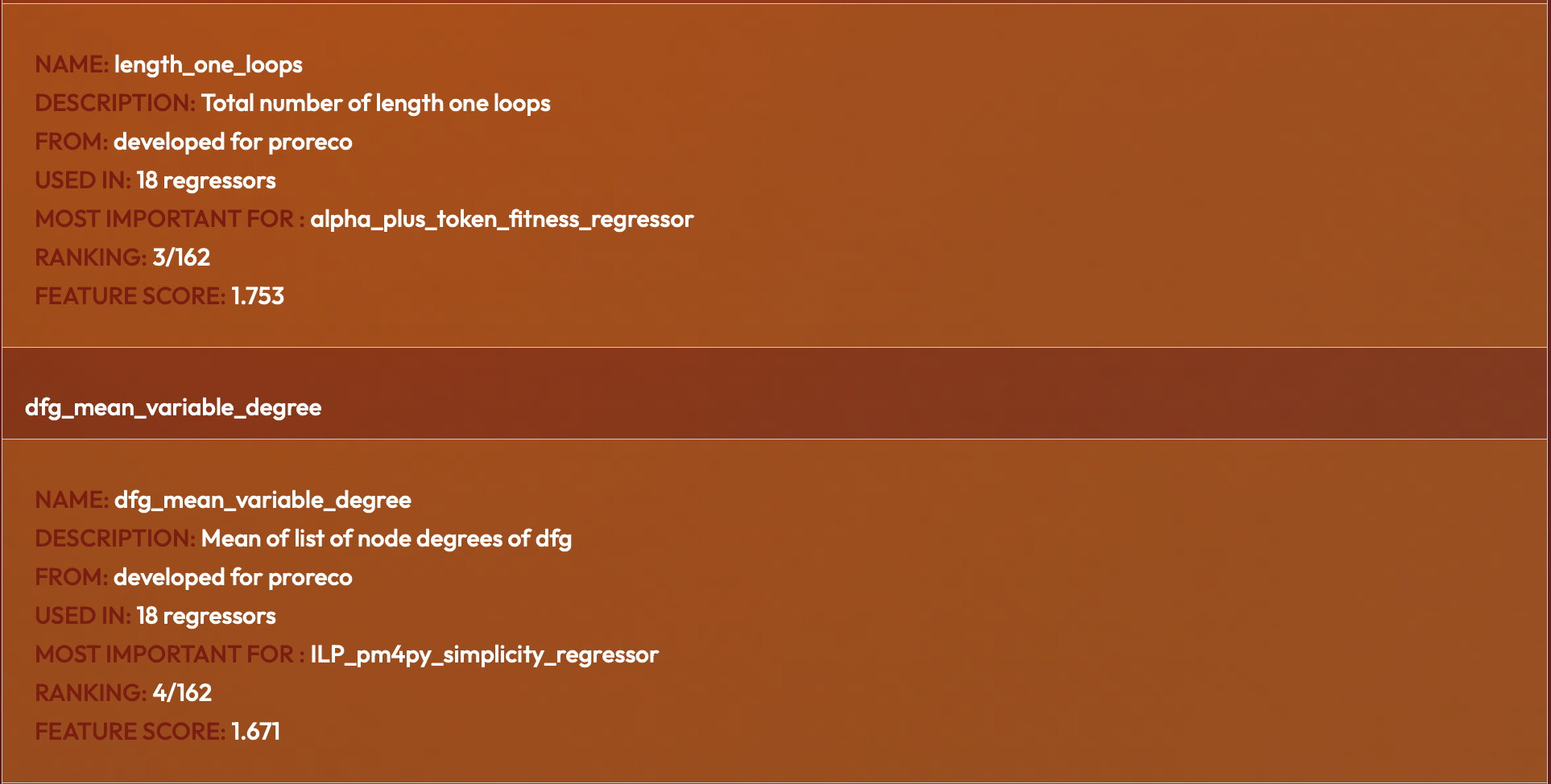}
    \caption{
        The user interface for the feature insights. 
    } \label{fig:feature-insights}
    % \vspace{-2em}
\end{figure}

\texttt{ProReco} offers insights into the features extracted from event logs. 
The \textit{"Featurer"} (shown in Fig.~\ref{fig:feature-insights}) is accessible through the navigation bar. 
\textit{Featurer} provides the user with detailed information about the features. 
By searching for a specific feature name, users can access the following information, as shown in Fig.~\ref{fig:feature-insights}:
\begin{itemize}
    \item Description: a brief description of the feature.
    \item From: the source of the feature.
    \item Used in: the number of regressors that use this feature.
    \item Most important for: the regressor that gains the most advantage from the feature.
    \item Ranking: the importance of the feature among all features.
    \item Feature Score: a metric used to determine the feature’s ranking.
\end{itemize}

\begin{figure}[h!]
    \centering
    \includegraphics[width=0.9\linewidth]{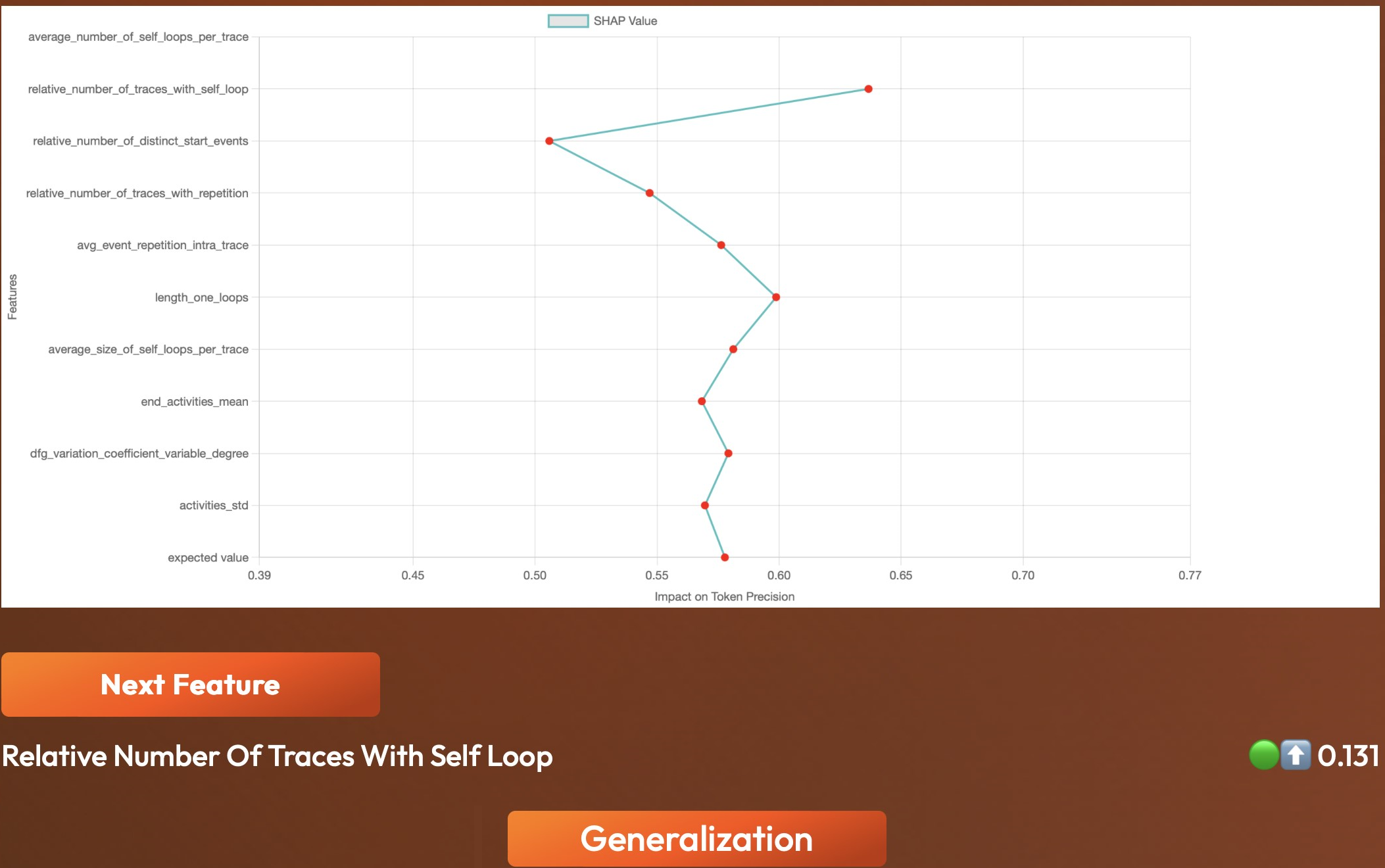}
    \caption{
        The user interface for explaining an individual recommendation.
    } \label{fig:feature-exaplanation}
    \vspace{-2em}
\end{figure}

\subsubsection{Explainable Recommendation}
As recommendations without explanations can hinder the transparency, trustworthiness, and satisfaction of a recommender system~\cite{ZhangC20XRecommendation}, \texttt{ProReco} incorporates techniques from explainable AI (XAI) to provide explanations for individual predictions made by the regressors. 
Users can access the explanations by clicking on “Explain the Predictions” on the resulting recommendation page. 
Then, users will be redirected to an interactive plot based on SHAP values~\cite{LundbergL17SHAP}. 
The plot begins at the bottom, displaying the expected measure for the selected algorithm. 
As each feature is added, its effect on the prediction is shown. 
A shift to the left indicates a decrease in the measure, while a shift to the right indicates an increase. 
The interactive plot offers insights and explanations for the recommendations.

%% file: sections/conclusion.tex
\section{Conclusion and Future Work}\label{sec:conclusion}
% \vspace{-0.5em}
Process discovery aims to automatically generate process models representing the underlying information system from event logs. 
Despite the development of various discovery algorithms and quality metrics, no single algorithm dominates in terms of model quality measures~\cite{AugustoCDRMMMS19PDreviewbenchmark}. 
Consequently, users often face the challenge of manually selecting suitable discovery algorithms, a process that is time-consuming and error-prone, even for experts in process mining. 
In response, this paper introduces \texttt{ProReco}, a process discovery recommender system designed to recommend the most appropriate process discovery algorithm based on user preferences and event log characteristics. 
ProReco expands upon previous work by incorporating state-of-the-art algorithms and providing transparent explanations for its recommendations.

As future work, several directions can be investigated. 
First, we plan to explore various parameter settings for the discovery algorithms. 
Due to the vast search space, the challenge is finding an efficient method to determine the best value per technique. 
Additionally, we would like to include additional measures for optimization such as runtime. 
Last but not least, we plan to conduct user studies to understand the effectiveness and usability of \texttt{ProReco} as well as to validate the benefits. 
For example, the understandability of the provided explanation could be evaluated by a dedicated user study. 